# Synthetic Medical Imaging Generation with Generative Adversarial Networks For Plain Radiographs

**Approved for Public Release; Distribution Unlimited. Public Release Case Number 22-3965**

**John R. McNulty[1]\*, Lee Kho[1], Alexandria L. Case[1], Charlie Fornaca[1], Drew Johnston[1], David Slater PhD[1], Joshua M. Abzug, MD[2], Dr. Sybil A. Russell, MD MPH[1]**

[1]      The MITRE Corporation, McLean, VA 22102, USA
[2]      University of Maryland School of Medicine, Departments of Orthopedics and Pediatrics, Baltimore, MD, 21201
\*      Author to whom correspondence should be addressed;

## Figures

## Tables

# Abstract

In medical imaging, access to data is commonly limited due to patient privacy restrictions and the issue that it can be difficult to acquire enough data in the case of rare diseases.[1] The purpose of this investigation was to develop a reusable open-source synthetic image generation pipeline, the GAN Image Synthesis Tool (GIST), that is easy to use as well as easy to deploy. The pipeline helps to improve and standardize AI algorithms in the digital health space by generating high quality synthetic image data that is not linked to specific patients. Its image generation capabilities include the ability to generate imaging of pathologies or injuries with low incidence rates. This improvement of digital health AI algorithms could improve diagnostic accuracy, aid in patient care, decrease medicolegal claims, and ultimately decrease the overall cost of healthcare.  The pipeline builds on existing Generative Adversarial Networks (GANs) algorithms, and preprocessing and evaluation steps were included for completeness.  For this work, we focused on ensuring the pipeline supports radiography, with a focus on synthetic knee and elbow x-ray images. In designing the pipeline, we evaluated the performance of current GAN architectures, studying the performance on available x-ray data. We show that the pipeline is capable of generating high quality and clinically relevant images based on a lay person's evaluation and the Fréchet Inception Distance (FID) metric.

# 1 Introduction

The purpose of this investigation was to address data shortages and patient privacy concerns in the medical imaging space by generating synthetic imaging data, specifically knee and elbow radiographs. Ultimately, this work provides groundwork towards a standardized methodology for generating clinically realistic synthetic medical images on a large scale in order to increase access to data and improve clinical algorithms.

# 2 Background

**2.1 Research Problem**

Though there are recognized benefits of artificial intelligence (AI) algorithms in the healthcare space, they require robust amounts of training data in order to be useful. Access to this data can be hindered by patient privacy protections. Furthermore, rarity of certain pathologies of interest also add to data scarcity. The combination of the aforementioned limitations greatly decreases the participation of smaller entities and independent researchers in the development of healthcare AI compared to larger institutions and/or medical systems who have data access advantages. These hindrances create timeline delays and are prohibitive to rapid development across the domain in clinical AI.

Data access can be streamlined through partnerships between developers and specific data providers, such as a hospital and/or medical system. However, the available data is often subpopulation specific. This results in siloed algorithm development where the algorithms are only applicable to a small population and can be difficult to scale for universal purposes. Furthermore, this creates difficulties comparing and regulating algorithms, particularly across geographic entities and different healthcare systems. These hurdles in access to data, lead to delays to project timelines and limitations to innovation relating to these data. Given the substantial cost and time barriers to access, the innovation in clinical AI has lagged compared to other fields.

The majority of research in general AI has been targeted toward creating larger and more complex models. However, with this increase in model size, access to larger amounts of data becomes more important than model architecture [2]. Additionally, when the size of the data used is large enough, simpler models have comparable performance to robust architectures [3]. This highlights a significant need for large amounts of training data. In fact, current predictions indicate that synthetic data will make up most data used for training models by 2030 [4]. This prediction illustrates the case for processes to easily supplement AI practitioners with useful synthetic data, as will be provided with the proposed pipeline.

In addition to a general need for synthetic data, there is also an increasing need for standardizing the evaluation of healthcare-related AI models. There is currently no agreed-upon standard evaluation metric for generative models [5]. The current pipeline supports these challenges by offering a standardized method for generating large datasets of synthetic medical images. In turn, this provides a pathway toward

"gold standard datasets" which can be used to compare performance between algorithms developed on disparate populations, as well as provide insights to shape the regulatory process of AI models for medical imaging.

## 2.2 Related Work

To remedy the lack of publicly available medical images, researchers have experimented with various frameworks for generating medical images over the years. According to a 2020 study, the most popular frameworks for generating synthetic medical data have been autoencoders, U-nets, and Generative Adversarial Networks (GANs)[6]. Recently U-nets and GANs have become the most common approaches to this task, with the popularity of GANs outpacing that of U-nets [6]. In addition to the availability of data, there is emerging research in creating GAN architectures designed for more complex imaging modalities, such as Computed Tomography (CT) scans [7].

Researchers in both academia and industry have expressed a strong interest in progressing medical AI, especially in the domain of radiology, as it is a popular clinical application of AI [8]. Specifically, radiographs of the pediatric elbow can create confusion due to the abundance of growth plates present on pediatric elbow radiographs. As such, injuries may be missed leading to delays in care, suboptimal outcomes, and medicolegal consequences, with the end result being an ultimate increase in healthcare costs. Google recently published a blog about the use of GANs in dermatologic image data [9], [10]. The National Institutes for Health (NIH) publicly released medical image datasets in 2018, and Stanford University releases similar datasets to the public on an ongoing basis [11], [12].

# 3 Methodology

## 3.1 Generative Adversarial Networks (GANs)

A generative adversarial network (GAN) is a state-of-the-art deep learning generative model that is comprised of two competing neural networks. The first one is a generative network that seeks to create increasingly realistic-looking data based on real data. In the case of the current study, the real data and the data that the generative network is working to emulate are medical images, specifically knee and elbow x-rays. The second network that comprises the GAN is a discriminative network that attempts to discern whether the generated data is real by comparing it with real data [13].

Training starts when the generator presents random noise to the discriminator, which is also presented with real data separately. The discriminator then tries to determine which data is real and which data is synthetic. The error from the discriminator is then backpropagated to the generator and the discriminator. As a result, both networks ultimately improve. Training continues until convergence is reached and either the generator stops making improvements or the discriminator can no longer discriminate between the images created by the generator and the training data.

## 3.2 Selecting a GAN

The current project surveyed several GAN architectures (including SinGAN, DCGAN, AnycostGAN, GIREAFFE, and IMAGINE) that have been recently developed [14], [15], [16], [17], [18]. Ultimately, the current project utilized StyleGAN3 for the pipeline due to its superior performance and ease of use [19]. StyleGAN3's ease of use was considered greater because of the availability of its code repository, wider adoption, and community support. The high image quality of its output also contributed to why StyleGAN3 was utilized. StyleGAN3 was also one of the most recently published architectures at the time of this investigation. By using StyleGAN3, high-quality synthetic images trained on various medical image datasets were successfully produced.

StyleGAN3 was developed by a research team at NVIDIA in late 2021 and at the time of this writing is the iteration of the StyleGAN family of frameworks. The model is implemented in PyTorch and has the ability to train a network on single- and multi-class datasets. It improves upon past versions of StyleGAN (such as StyleGAN2, StyleGAN, and Progressive Growing of GANS) and leverages the past success of those architectures including a style-based generator (which is a more complex architecture that allows for scale-specific control of image generation), the elimination of various artifacts, and the ability to stabilize training with limited data (e.g. 1000's of images) [19].

## 3.3 Pipeline

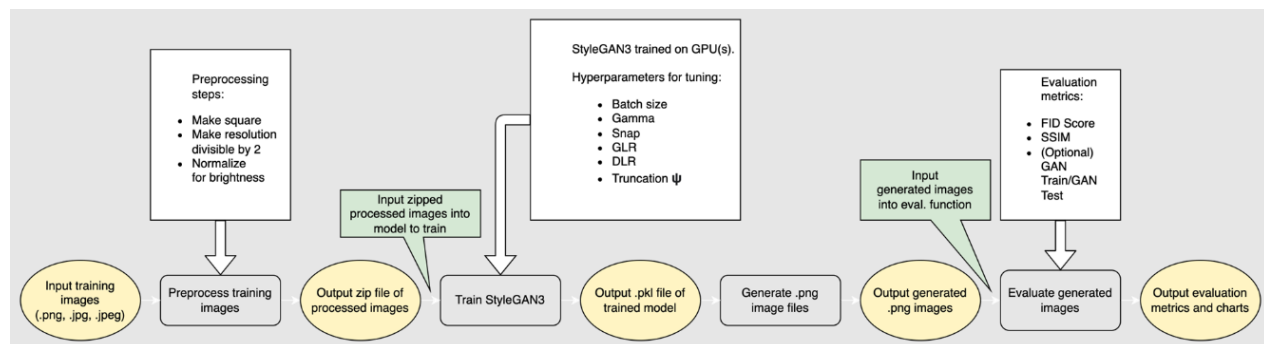

*Figure 1: Pipeline Diagram*

### 3.3.1 Diagram/Overview

The purpose of the pipeline is to make medical image generation with GANs easier to use for researchers and society. The pipeline was designed to be an "end-to-end" to ensure ease of use. As such, the pipeline incorporates every step of the modeling process, which includes data preprocessing, training of the GAN, image generation, and evaluation of the synthetic images.

The model also allows users to customize certain parameters to suit their individual purposes. One aspect they can set is the portion of the pipeline that they wish to run. For example, a user can indicate that they want to run the preprocessing and training steps and forego the generation and evaluation steps. In addition, there are several hyperparameters specific to StyleGAN3 that the user may experiment with to achieve the highest quality results for their dataset. These hyperparameters are detailed in the StyleGAN3 repository documentation.

To easily create a container environment for the pipeline with the necessary packages, a Docker file in the repository was included. It is important to note that the pipeline requires access to at least one GPU because the training process can be computationally intensive. To that end, the pipeline is platform agnostic. In other words, it can be run via local GPU, remote GPU, job scheduler, etc. This means that users are responsible for any scripts/configurations necessary to run the pipeline on their GPU platform(s). As an example, the documentation includes instructions for running the pipeline in a SLURM environment.

### 3.3.2 Preprocessing

StyleGAN3 has certain requirements of the image data which must be met in order to run training. These requirements include ensuring that the image data is formatted to be in the shape of a square, has a resolution that is a power of two, and the data is contained within in a zipped file. Additionally, any image labels must be contained in a separate JSON file. The GIST preprocessing script includes functions to help the user easily ensure that these requirements are met before training begins. If needed, the user also has the option to split the data into a train/test set.

### 3.3.3 Training and Generation

The Stylegan3 repository includes sufficient functionality for training a model and generating images from that model. As such, we leverage that functionality directly and include it in the pipeline with minimal to no changes. Stylegan3 takes in the dataset and trains a GAN model as described in section 3.1. In addition to training directly from data, users also have the option to use provided pretrained networks for transfer learning and continued training. All training runs are saved as pickled models. The StyleGAN3 repository includes hyperparamters which can be used to control how Stylegan3 trains on the data provided. The default hyperparameters (such as generator learning rate, discriminator learning rate, and batch size) have been optimized to work for the majority of cases. The repository's readme file includes general guidelines and recommendations on which configurations are best to start with, as well as which direction to tune the hyperparameter values based on performance. The current investigation has found that the default configurations were sufficient. Additionally, it was observed that hyperparameter tuning did not lead to noticeable improvement in performance over the recommended starting values.

The StyleGAN3 repository also includes a script for generating synthetic images based on a previously trained model. Generating synthetic images can be an important component of initial experimentation, as it allows for manual inspection of the images. Additionally, this enables the user to create datasets of synthetic images from their trained models.

### 3.3.4 Evaluation (FID score, GAN-train/GAN-test)

The Stylegan3 repository includes multiple quantitative metrics for evaluating performance such as FID score, precision, recall, and equivariance, for evaluation of image quality. The FID score is widely accepted for evaluating GAN performance [20]. The FID score is calculated by comparing the generated images distribution with the training image distribution and can be re-calculated for every exported model.

These quantitative metrics are helpful in terms of evaluation, but they do not capture all the issues found in the resulting synthetic images. These unrealistic issues present in the synthetic images were defined as "artifacts". Examples of artifacts found, such as improper joint formation, x-ray blurriness, and bone curvature, were found as this investigation experimented with training. These examples are shown in section 4.2. It is recommended to perform a qualitative check on the synthetic images, such as a manual inspection for artifacts, in addition to a quantitative check, in order to create a more well-rounded understanding of performance.

Because the quantitative metrics provided by Stylegan3 did not capture these artifacts, another quantitative approach from the literature was added to GIST. This approach, known as the "Gan-train/Gan-test" approach, is based on the work of Shmelkov et. Al [21]. "Gan-train/Gan-test" calculates a score for two machine learning classifiers. The first is trained on GAN-generated synthetic images and then tested on a holdout set of real images. The second approach is trained on real images and tested on synthetic images. The resulting metrics, such as precision and recall, indicate the performance of the model.

## 3.4 Data

Various publicly available x-ray datasets were used in the construction and testing of the pipeline. These datasets are listed in Table 1. The datasets ranged in image complexity. For example, the KneeXrayOA-simple dataset ensured each image contained the exact same view of the knee and that the knee was centered in the image, filling the entirety of the space [22]. Conversely, the MURA elbow dataset was more complex with straight, lateral, and oblique views for every elbow [23]. Furthermore, the x-ray was not always centered in the image and would often contain additional objects, such as a radiologist tag. These additional complexities made it more challenging for the GAN to achieve similar model performance across the two datasets.

Additional complexity was present in the University of Maryland elbow radiographs. This dataset was collected specifically for the purposes of this study and was approved by an Institutional Review Board from the University of Maryland School of Medicine. These radiographs were of the pediatric elbow and thus had the presence of various growth plates, elbow in different stages of development/ossification, and a lack of standardized rotation on the views. The attempted benefit of this dataset was the realistic data that, in theory, could utilize AI technology in the future.

| Name of Dataset | Location | Type of Data | Image Count | Shortcomings/Challenges |
|---|---|---|---|---|
| **KneeXrayOA-simple** | Kaggle | Osteoarthritic Knee X-Rays (JPG) | 10k | -Similarity of images<br>-Lack of labels (e.g.: medical implants)<br>-Light saturation |
| **MURA** | Stanford | Musculoskeletal Radiographs (PNGs) | 5k (elbow) | -Inconsistent positioning of x-ray<br>-Lack of labels |
| **UMD Elbow** | University of Maryland | Elbow X-Rays (DICOM) | < 1k | - Dataset Size<br>- Inconsistent positioning of X-ray |

*Table 1: Datasets used*

## 3.5 Experimentation

### 3.5.1 Initial Experimentation

For each dataset, the GIST pipeline was utilized to train models, generate synthetic images, and evaluate the performance of the GAN. While the platform is agnostic to the GPU environment used, the pipeline ran on a high-performance compute environment using SLURM, an open-source job scheduler for Linux and Unix-like kernels.

An example training process for the KneeXrayOA dataset can be seen in Figure 2. The generator begins by presenting a random collection of pixels to the discriminator. As the error is backpropagated, the model improves, and one begins to see the x-ray of the knee form. Training continues to progress until convergence is reached and a fairly realistic knee x-ray remains.

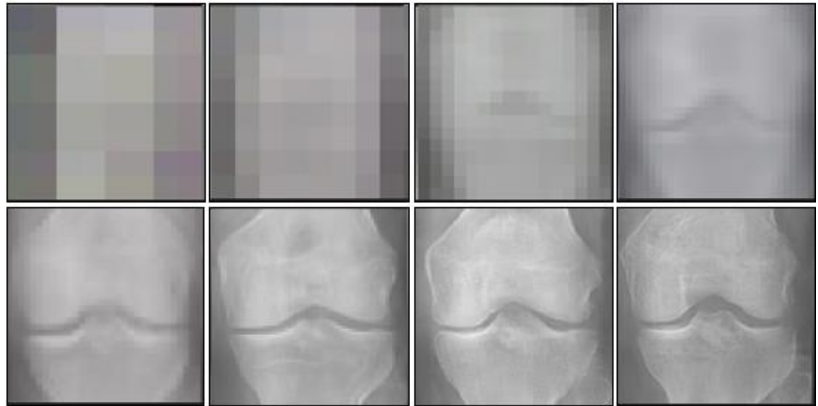
*Figure 2: GAN training progression*

As training progressed, the fid50_full score was leveraged, which calculated the FID score for the current model against the training set of images. Eventually, convergence was reached. In addition to observing the fid50_full score, synthetic images from the models at the point of convergence were generated and manually inspected for artifacts. Depending on model performance, the hyperparameters were tuned, data was added, and data augmentation was performed in attempts to improve the model performance.

While the two public datasets used in this study had sufficient data for training, medical image datasets are often small in size due to the issues previously mentioned. Therefore, future users of the pipeline may not have access to as much data for training. As such, an investigation was performed to determine if a threshold (minimal) amount of data was required for convergence of the fid50_full score, and a lack of artifacts. In other words, what minimal training dataset size is required for successful image generation. This could inform future users if they had enough data to pursue use of the pipeline. To establish this threshold, the same experiment was run for various training dataset sizes and the performance was evaluated. Due to the use of the PyTorch package, it was not guaranteed that two training runs, trained on the same dataset with the same hyperparameter values, would converge to the exact same result. This implies that a direct comparison between runs may include slightly different results; an implication which was validated by the later analysis. However, the overall comparison is still useful as the variability between the same runs was less than the difference between runs with varying amounts of data, or different hyperparameters.

### 3.5.2 Hyperparameter Tuning

In deep learning, hyperparameter tuning tests the adjustment of model architecture and training changes, and their impact on performance. Due to the structure of GAN architecture, and the competition between the two neural networks, GANs are particularly sensitive to hyperparameter tuning. Specifically, the parameters are at risk of oscillating and becoming unstable. Therefore, hyperparameter tuning can require much effort and experimentation. Of the hyperparameters present in the StyleGAN3 model, the authors suggest beginning tuning with the most impactful hyperparameters first. Based on author suggestion, the hyperparameters that were turned included batch size, gamma (R1 regularization), and discriminator learning rate [19].

### 3.5.3 Threshold Data Requirements

Increased image complexity can include greater variability within the x-ray image, such as the view or angle of a joint. It can also include a greater number of bones/ossification centers and organs present in the image. In order to ensure that any models trained on this data have sufficient ability to capture this complexity, a larger image count in the training data is required. Due to this need for additional data, the current study investigated if a minimum threshold amount of data could be established to achieve quality synthetic x-rays. Different training dataset sizes were used from the MURA elbow dataset and performance was evaluated via the FID score, as well as, manual inspection for artifacts. Six different training runs were conducted using 100, 200, 300, 400, 500, and 781 images, respectively. The largest dataset of 781 images represented the total count of right elbow lateral view x-rays from the MURA elbow dataset. The dataset size was used for the other experiments mentioned above.

In addition to training base models on different dataset sizes, the effect of dataset size on transfer learning was also explored. A base model trained on the MURA lateral elbow dataset was trained until it reached convergence. The model was then used for transfer learning where the new training was built upon the base model but uses the varying dataset sizes to continue training.

### 3.5.4 Clinician Evaluation

In order to further evaluate the performance of the pipeline on x-rays, a "blind test" was conducted. 50 images from the KneeXrayOA Knee training dataset were selected, along with 50 synthetic generated knee images. These images were re-labelled and randomly shuffled. A "ground truth" dataset of 20 images was also selected from the training set to be a representative sample of the distribution of classes found in the overall dataset. The "ground truth" dataset was meant to orient the participant, an orthopedist with ample experience interpreting x-rays, on the appearance of knee x-rays from this dataset. The participant then labelled each image in the blind set as either "real" or "synthetic".

### 3.5.5 Gan-train/Gan-test

The knee osteoarthritic dataset was used to test the performance of the "Evaluate" section of the pipeline. A model was trained that differentiated between the five classes of x-ray image and was saved at the point of convergence according to the FID score. The evaluate section of the pipeline was then run using this model, and performance metrics were recorded. For context, these classes are similar enough in appearance that a layperson would also find differentiation challenging. So then manual analysis was then performed looking solely at two more differentiated classes, class 0 (lowest level of osteoarthritis) and class 3 (second-highest level of osteoarthritis). The reason class 3 was selected over class 4 (highest level of osteoarthritis) is because of the low number of images present in class 4.

Three pretrained classification models from the torchvision models package were trained according to the "Gan-train/Gan-test" framework. These models included inception, resnet101, and vgg19 [24], [25], [26] Each model was trained according to a base implementation, meaning they were trained on real images, as well as, evaluated on real images, a "gan-train" implementation meaning they were trained on synthetic images and evaluated on real images, and a "gan-test" implementation meaning they were trained on real images and evaluated on synthetic images.

# 4 Results

### 4.1 Training Results: FID score convergence across x-ray types

The FID score is relative to the complexity of the data measured. As the complexity of the set of images increases, so will the FID score. Therefore, convergence at a lower FID score for one type of x-ray image does not necessarily indicate improvement over models for different x-rays that have converged at a higher score. For example, when training on the Knee-OA data, the model converged to a FID score of 65.23 while the model trained on the MURA data converged to a 75.89. Run count for training times varied among the different datasets based on both image complexity and the number of images used for training.

**4.2 Image Generation Results: Qualitative analysis of image artifacts**

Various artifacts were noted during the manual inspection of the generated x-ray images. Using the MURA elbow dataset, the various artifacts that were noted were due to the complexity of the differing views of the elbow. After subsetting the dataset to only the lateral view of the elbow, the artifacts underwent a noticeable improvement. The human elbow is a hinge joint (Figure 3, a1). We found that this particular joint type does not render particularly well when synthetically generated. However, after training using only the lateral view of the elbow joint, we saw the hinge portions begin to render more accurately to the true structure of the joint (Figure 3, a2). An unrealistic curvature of the bone was also seen in the ulna forearm bone (Figure 3, b1) which also improved (Figure 3, b2). The training dataset contained images of x-rays, where the x-ray was a much smaller portion of the overall image. In certain cases, the model did not render these smaller x-rays particularly well (Figure 3, c1) though improvement was seen on training of the lateral subset (Figure 3, c2).

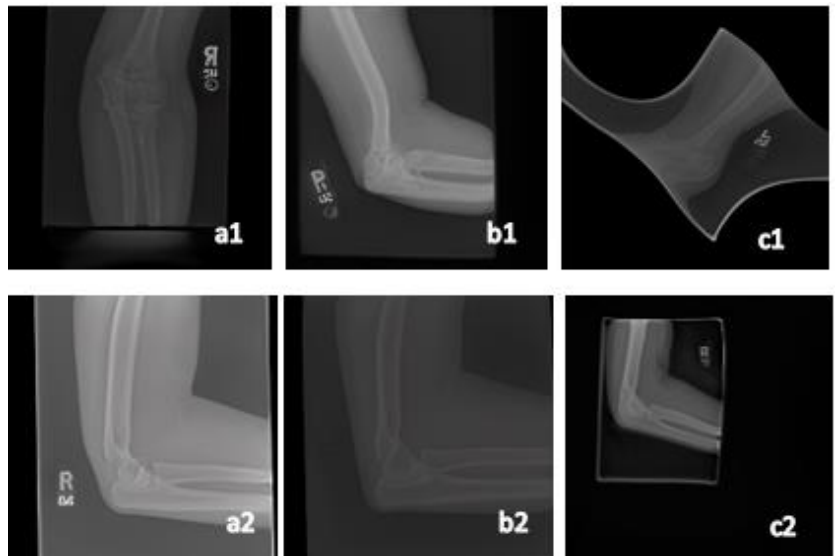
*Figure 3: Elbow artifacts*

**4.3 Hyperparameter Tuning**

After increasing and decreasing key hyperparameter values in varying combinations, it was discovered that tuning did not contribute significant improvements over the base model recommended by the authors. Results of the top performing combinations of hyperparameters are shown in Figure 4. While there was marginal improvement in some cases, it was not greater than the variability seen across the same runs and could also be a result of additional training time.

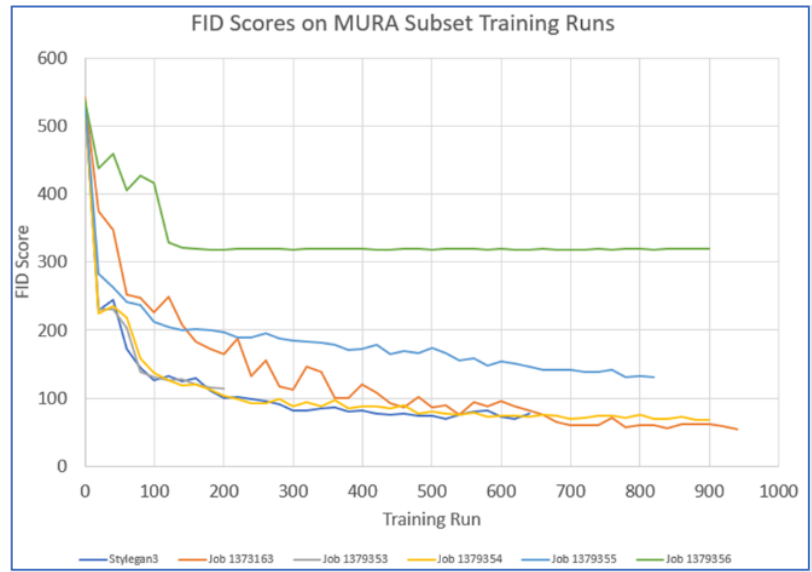
*Figure 4: Results of Hyperparameter Tuning*

**4.4 Training Dataset Size**

The results of training on varying dataset size according to FID score can be seen in Figure 5. FID score performance improved with increasing size of the training dataset, though notably after the dataset size increased to 200+ images, improvement in the FID score was fairly marginal with increasing dataset size. Images were generated during training at every 20$^{th}$ training iteration, and the generated images at the point of convergence were inspected for artifacts. The smaller training datasets (500 images or fewer) produced images with artifacts present. Some examples of these artifacts can be found in Figure 6. However, the run trained on 781 images produced no artifacts. Therefore, in this case the FID score alone was not sufficient for determining the quality of the images, as the smaller datasets converged to a similar (albeit larger) FID score than the largest dataset of 781 images. For lateral elbow x-rays, a threshold training dataset size of greater than 500 images is required to generate realistic, artifact-free images. It should be noted that the complexity of images will vary across the anatomy being x-rayed, as well as, the view of the x-ray. Thus, this threshold is meant to serve as a rough baseline for lateral elbow x-rays but should not be construed as applicable to any x-ray type.

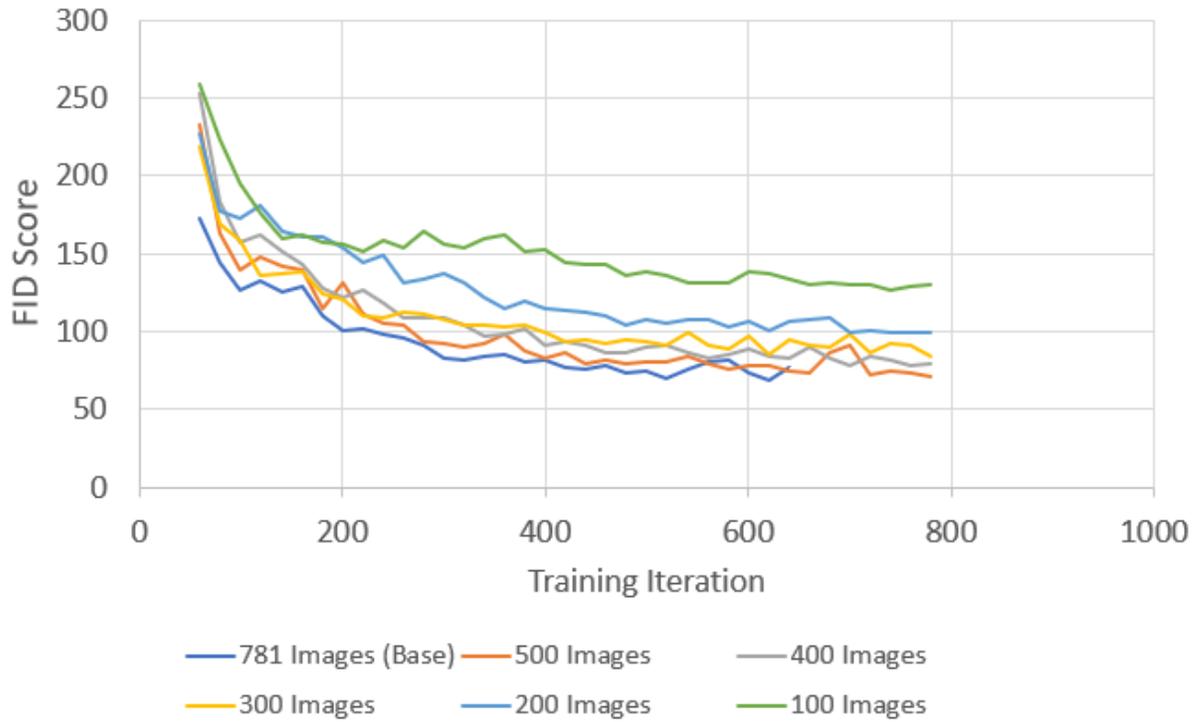

*Figure 5: FID Score Results of Varying Training Dataset Size*

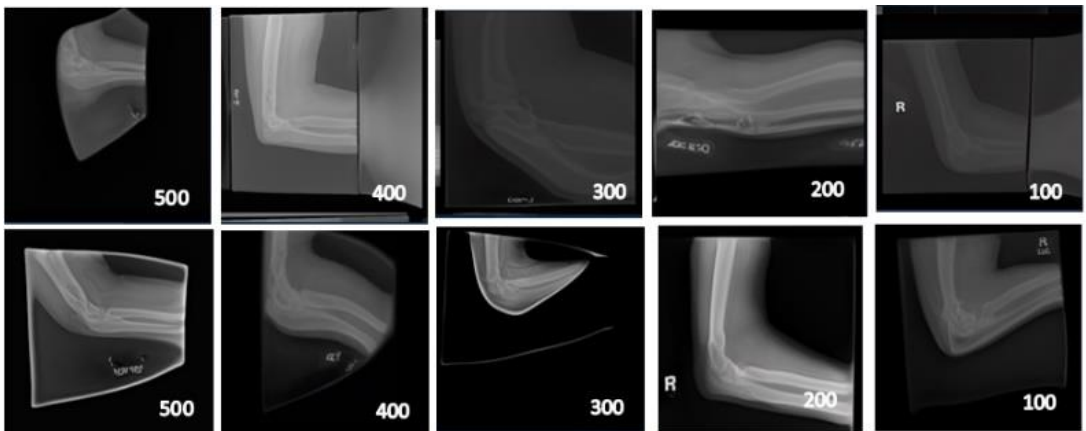

*Figure 6: Artifacts discovered when training on smaller datasets*

This analysis is further supported by the results from leveraging transfer learning. Performance of each model that continued training from the base model improved with dataset size. The results can be seen in Figure 7. In the figure we can see that each time a smaller dataset was used, the generated images included artifacts. This denotes inferior performance, especially when compared with the base model trained on 781 images, whose results can be seen in Figure 3 a2, b2, and c2. This suggests that in order to generate images free from artifacts, a model will need at least 750 images for training.

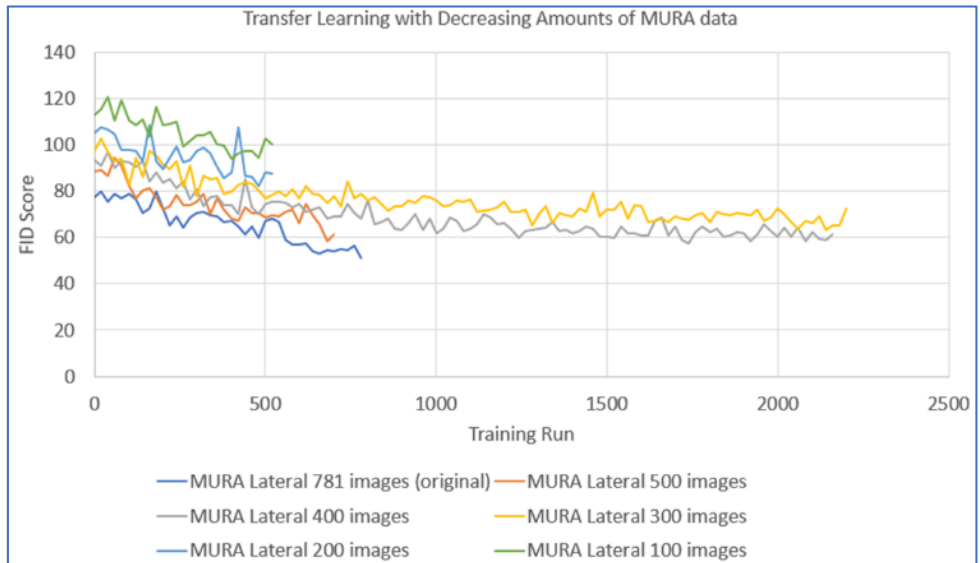
*Figure 7: Results of transfer learning using smaller datasets*

### 4.5 Clinician's Evaluation

The blind test resulted in an accuracy of 53%, a precision of 58%, and a recall of 58%. These results suggest that the participant could not differentiate between the real and synthetic images. This in turn suggests that the model performed well in generating realistic images.

### 4.6 Gan Train Gan Test Evaluation

The results of the Gan-train/Gan-test evaluation can be seen in Figure 10. Most of the "gan-train" and "gan-test" models performed worse according to accuracy when compared to the "base" model (the control) for each classification type. In the case of Gan Train, where the classifiers were trained on synthetic data and evaluated on real data, the accuracies were all 17.03%-18.53% different than the base model. There was greater variability in the Gan Test results, where the classifiers were trained on real data and evaluated on synthetic data. In this case, the Inception classifier performed best at only a 5.68% difference compared to the base accuracy, while Resnet101 (22.66% difference) and VGG19 (31.87% difference) performed worse. However, these models outperformed a dummy classifier which would have a 50% accuracy. This suggests that the synthetic images have predictive power for a classification model, though admittedly less than real images.

|  |  | Accuracy | Precision | Recall | F1 Score |
| --- | --- | --- | --- | --- | --- |
| Inception | Base | 78.36 | 78.43 | 78.37 | 78.35 |
|  | Gan Train | 65.87 | 68.78 | 65.87 | 64.49 |
|  | Gan Test | 74.03 | 76.56 | 74.04 | 73.41 |
| Resnet101 | Base | 82.69 | 83.00 | 82.69 | 82.65 |
|  | Gan Train | 69.71 | 73.77 | 69.71 | 68.36 |
|  | Gan Test | 65.86 | 69.94 | 65.86 | 64.02 |
| VGG19 | Base | 82.21 | 83.10 | 82.21 | 82.09 |
|  | Gan Train | 68.27 | 73.21 | 68.27 | 66.48 |
|  | Gan Test | 59.61 | 63.54 | 59.61 | 56.46 |

*Table 2: Evaluation Results*

# 5 Discussion

We have presented an open-source pipeline for the purposes of making the development and evaluation of synthetic medical imagery more accessible. To demonstrate the effectiveness of the pipeline, we worked with two radiology image datasets. When the pipeline's performance was evaluated, it was found that it performed well according to the FID score, manual inspection, and the blind test, suggesting that the pipeline can be used to make realistic medical imagery.

Some important lessons learned and suggested best practices discovered through this process should be noted. The amount and structure of classes in the training data makes a difference for model training performance. Model performance and quality increases with access to more data as shown in Figure 5 and Figure 6. Models tend to perform better when trained on distinct classes. For example, training a model on only the lateral view of the elbow, rather than every view treated as a single class, will improve performance. Additionally, while hyperparameter tuning can prove useful and increase performance, model architecture recommendations are often already optimized for best results.

**5.1 Further Research**

The quantitative analysis is not sufficient alone for the purpose of evaluating model performance. Metrics such as the fid score are generally very useful but did not capture the artifacts present in the generated images. Because of this, manual inspection was required. The field could be improved by expanding existing metrics and developing more robust new metrics that can easily catch artifacts. This would in turn decrease the need for human-in-the-loop inspection.

The fact that the blind test participant could not distinguish between "real" and "fake" images is significant as it indicates the images are realistic and therefore potentially useful. Since this pipeline was created with the intention of supporting clinical AI with synthetic datasets, the natural next step is to test the performance of machine learning classifiers trained on this synthetic data and tested on real data to see if the synthetic images are useful from a machine learning perspective. The "gan train gan test" portion of the pipeline is an initial step in this analysis.

It is the goal that clinicians and data science practitioners who are not experienced with deep learning can use the current investigation's pipeline to more easily create synthetic images for their clinical AI. While the current pipeline is limited to two-dimensional images, there is already research into architectures for generating three-dimensional images, like CT-scans, which could broaden the scope of a future pipeline.

**5.2 Further Development**

In addition to supplemental research, further development could also encourage adoption of the pipeline. The current pipeline is designed to work "off the shelf" for users, but requires at least a basic level of programming expertise to clone and use the repo. This could be a rate limiting step or put the pipeline out of reach for some potential users. To mitigate this burden, the coding requirement could be removed if the pipeline was housed as a web hosted product. Such a product would require a basic user interface, backend services to store the data and run the pipeline on a gpu instance. As a result, it may require a fee to cover the cost of hosting and using cloud-based services.

# 6 Conclusion

The pipeline in the current study was created to allow greater access to both clinicians and data scientists who are interested in creating realistic synthetic data but are not well-equipped enough in their technical expertise to make use of repositories like StyleGAN3 off the shelf. Developing that technical experience requires extensive training, time, and money to acquire. The current investigation's pipeline abstracts away these deep learning architectures for the user.

It is the goal that this pipeline can be used for the quick curation of synthetic datasets and can contribute to clinical AI research efforts as a result. Overall, the investigation has been successful at making a pipeline for practitioners to generate synthetic x-ray images for themselves. This is not limited to x-ray images, as any two-dimensional image will work.

The current study has also indicated potential for future work, to include the development of a set of more robust evaluation metrics, further testing of models trained or tested on synthetic data, as well as, research into more complex imaging modalities, such as CT scans and magnetic resonance imaging (MRI) images.

# 7 Acknowledgement

This research was funded by MITRE's Independent Research and Development Program.

# 8 References

[1] A. S. Coyner *et al.*, "Synthetic Medical Images for Robust, Privacy-Preserving Training of Artificial Intelligence," *Ophthalmol. Sci.*, vol. 2, no. 2, p. 100126, Feb. 2022, doi: 10.1016/j.xops.2022.100126.
[2] "The Unreasonable Effectiveness of Data | IEEE Journals & Magazine | IEEE Xplore." Accessed: Apr. 20, 2023. [Online]. Available: https://ieeexplore.ieee.org/document/4804817
[3] I. Tolstikhin *et al.*, "MLP-Mixer: An all-MLP Architecture for Vision." arXiv, Jun. 11, 2021. doi: 10.48550/arXiv.2105.01601.
[4] L. Ramos and J. Subramanyam, "Maverick* Research: Forget About Your Real Data — Synthetic Data Is the Future of AI," Gartner, Inc, Jun. 2021. [Online]. Available: https://www.gartner.com/document/4002912
[5] A. Alaa, B. V. Breugel, E. S. Saveliev, and M. van der Schaar, "How Faithful is your Synthetic Data? Sample-level Metrics for Evaluating and Auditing Generative Models," in *Proceedings of the 39th International Conference on Machine Learning*, PMLR, Jun. 2022, pp. 290–306. Accessed: Apr. 20, 2023. [Online]. Available: https://proceedings.mlr.press/v162/alaa22a.html
[6] T. Wang *et al.*, "A review on medical imaging synthesis using deep learning and its clinical applications," *J. Appl. Clin. Med. Phys.*, vol. 22, no. 1, pp. 11–36, Jan. 2021, doi: 10.1002/acm2.13121.
[7] Y. Mirsky, T. Mahler, I. Shelef, and Y. Elovici, "CT-GAN: Malicious Tampering of 3D Medical Imagery using Deep Learning".
[8] "Top 5 Use Cases for Artificial Intelligence in Medical Imaging." Accessed: Apr. 20, 2023. [Online]. Available: https://healthitanalytics.com/news/top-5-use-cases-for-artificial-intelligence-in-medical-imaging
[9] "Generating Diverse Synthetic Medical Image Data for Training Machine Learning Models – Google AI Blog." Accessed: Apr. 20, 2023. [Online]. Available: https://ai.googleblog.com/2020/02/generating-diverse-synthetic-medical.html

[10]  A. Gohorbani, V. Natarajan, D. D. Coz, and Y. Liu, "DermGAN: Synthetic Generation of Clinical Skin Images with Pathology," 2019. Accessed: Apr. 20, 2023. [Online]. Available: https://arxiv.org/abs/1911.08716

[11]  "Shared Datasets | Center for Artificial Intelligence in Medicine & Imaging." Accessed: Apr. 20, 2023. [Online]. Available: https://aimi.stanford.edu/shared-datasets

[12]  "NIH Clinical Center releases dataset of 32,000 CT images | National Institutes of Health (NIH)." Accessed: Apr. 20, 2023. [Online]. Available: https://www.nih.gov/news-events/news-releases/nih-clinical-center-releases-dataset-32000-ct-images

[13]  I. Goodfellow *et al.*, "Generative Adversarial Nets," in *Advances in Neural Information Processing Systems*, Curran Associates, Inc., 2014. Accessed: Apr. 20, 2023. [Online]. Available: https://papers.nips.cc/paper_files/paper/2014/hash/5ca3e9b122f61f8f06494c97b1afccf3-Abstract.html

[14]  T. R. Shaham, T. Dekel, and T. Michaeli, "SinGAN: Learning a Generative Model from a Single Natural Image." arXiv, Sep. 04, 2019. Accessed: Aug. 10, 2023. [Online]. Available: http://arxiv.org/abs/1905.01164

[15]  A. Radford, L. Metz, and S. Chintala, "Unsupervised Representation Learning with Deep Convolutional Generative Adversarial Networks." arXiv, Jan. 07, 2016. doi: 10.48550/arXiv.1511.06434.

[16]  J. Lin, R. Zhang, F. Ganz, S. Han, and J. Zhu, "Anycost Gans for interactive image synthesis and editing," *arXiv.org*, [Online]. Available: https://arxiv.org/abs/2103.03243

[17]  M. Niemeyer and A. Geiger, "GIRAFFE: Representing Scenes as Compositional Generative Neural Feature Fields." arXiv, Apr. 29, 2021. doi: 10.48550/arXiv.2011.12100.

[18]  P. Wang, Y. Li, K. K. Singh, J. Lu, and N. Vasconcelos, "IMAGINE: Image Synthesis by Image-Guided Model Inversion." arXiv, Apr. 12, 2021. doi: 10.48550/arXiv.2104.05895.

[19]  T. Karras *et al.*, "Alias-Free Generative Adversarial Networks".

[20]  M. Heusel, H. Ramsauer, T. Unterthiner, B. Nessler, and S. Hochreiter, "GANs Trained by a Two Time-Scale Update Rule Converge to a Local Nash Equilibrium," in *Advances in Neural Information Processing Systems*, Curran Associates, Inc., 2017. Accessed: Apr. 24, 2023. [Online]. Available: https://papers.nips.cc/paper_files/paper/2017/hash/8a1d694707eb0fefe65871369074926d-Abstract.html

[21]  K. Shmelkov, C. Schmid, and K. Alahari, "How Good Is My GAN?," in *Computer Vision – ECCV 2018: 15th European Conference, Munich, Germany, September 8-14, 2018, Proceedings, Part II*, Berlin, Heidelberg: Springer-Verlag, Sep. 2018, pp. 218–234. doi: 10.1007/978-3-030-01216-8_14.

[22]  "KneeXrayOA-simple." Accessed: Apr. 25, 2023. [Online]. Available: https://www.kaggle.com/datasets/tommyngx/kneexrayoa-simple

[23]  "MURA Dataset: Towards Radiologist-Level Abnormality Detection in Musculoskeletal Radiographs." Accessed: Aug. 10, 2023. [Online]. Available: https://stanfordmlgroup.github.io/competitions/mura/

[24]  "Inception_v3," PyTorch. Accessed: Dec. 17, 2023. [Online]. Available: https://pytorch.org/hub/pytorch_vision_inception_v3/

[25]  "resnet101 — Torchvision main documentation." Accessed: Dec. 17, 2023. [Online]. Available: https://pytorch.org/vision/main/models/generated/torchvision.models.resnet101.html

[26]  "vgg19 — Torchvision main documentation." Accessed: Dec. 17, 2023. [Online]. Available: https://pytorch.org/vision/main/models/generated/torchvision.models.vgg19.html